\documentclass[final,5p,times,twocolumn]{elsarticle}

\usepackage{graphicx}
\usepackage{amssymb}
\usepackage{lineno}

\usepackage{mathtools} % for equations

\usepackage{multirow} % for table multirow

\usepackage[inline]{enumitem}

\usepackage{color}

\journal{}

\begin{document}

\begin{frontmatter}


\title{A Deep Learning Approach to Detect Lean Blowout in Combustion Systems}

% \author{Name\corref{cor1}\fnref{label2}}
% \cortext[cor1]{Corresponding Author}


\author[1]{Tryambak Gangopadhyay\fnref{label1}}
\author[2]{Somnath De\fnref{label2}}
\author[3]{Qisai Liu}
\author[2]{Achintya Mukhopadhyay}
\author[2]{Swarnendu Sen}
\author[3]{Soumik Sarkar\corref{cor1}}

\fntext[label1]{Work performed while the author was with Iowa State University prior to joining Amazon.}
\fntext[label2]{Present Address: Department of Aerospace Engineering, Indian Institute of Technology Madras, Chennai, India, 600036.}
\cortext[cor1]{To whom correspondence may be addressed. Email: soumiks@iastate.edu.}

\address[1]{Amazon Web Services AI, Amazon, Santa Clara, CA, USA.}
\address[3]{Department of Mechanical Engineering, Iowa State University, Ames, IA, USA}
\address[2]{Department of Mechanical Engineering, Jadavpur University, Kolkata, India}

\begin{abstract}
Lean combustion is environment friendly with low $NO_{X}$ emissions and also provides better fuel efficiency in a combustion system. However, approaching towards lean combustion can make engines more susceptible to lean blowout. Lean blowout (LBO) is an undesirable phenomenon that can cause sudden flame extinction leading to sudden loss of power. During the design stage, it is quite challenging for the scientists to accurately determine the optimal operating limits to avoid sudden LBO occurrence. Therefore, it is crucial to develop accurate and computationally tractable frameworks for online LBO detection in low $NO_{X}$ emission engines. To the best of our knowledge, for the first time, we propose a deep learning approach to detect lean blowout in combustion systems. In this work, we utilize a laboratory-scale combustor to collect data for different protocols. We start far from LBO for each protocol and gradually move towards the LBO regime, capturing a quasi-static time series dataset at each condition. Using one of the protocols in our dataset as the reference protocol and with conditions annotated by domain experts, we find a transition state metric for our trained deep learning model to detect LBO in the other test protocols. We find that our proposed approach is more accurate and computationally faster than other baseline models to detect the transitions to LBO. Therefore, we recommend this method for real-time performance monitoring in lean combustion engines.
\end{abstract}


\end{frontmatter}

% \linenumbers

\section{Introduction}

Nowadays, one of the primary technological problems of fuel-rich combustors in industrial and rocket propulsion applications is $NO_{X}$ emissions, which are influenced by local high flame temperatures \cite{zeldvich1946oxidation}.
The necessity to have low $NO_{X}$ emissions motivates scientists to move towards lean combustion \cite{correa1993review, mongia1998aero,turns1996introduction, mukhopadhyay2019fundamentals}.
Lean combustion not only reduces emissions but also enhances combustion efficiency with better fuel utilization. Therefore, such an approach is aimed towards a cleaner and better environment. However, during such ultra-lean operation regimes, the fuel-air ratio becomes very low, making the flame unstable, and the flame can even get extinguished by blowing out of the combustion chamber.
Such an unwanted event in the operation of combustion systems is referred to as lean blowout (LBO)~\cite{muruganandam2005active,gupta2019prevention,de2019use,de2020application}. The chances of LBO get more pronounced when the unburnt mixture velocity gets higher than the reacting speed \cite{turns1996introduction}.

\begin{figure*}
\centering
\includegraphics[width=14cm, keepaspectratio]{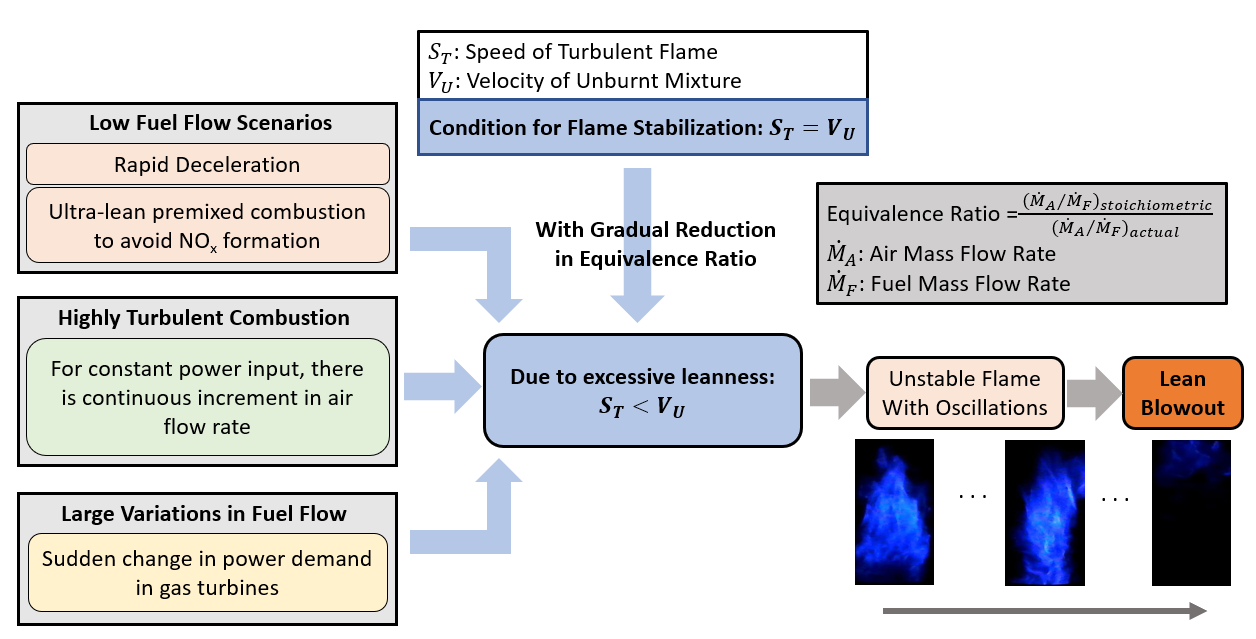}
\caption{Illustration of the concept of lean blowout (LBO) in combustion systems.}
\label{lbo_concept} 
\end{figure*}

LBO may occur during scenarios like low fuel flow, highly turbulent combustion, and large variations in fuel flow.
The concept of LBO in illustrated in Fig.~\ref{lbo_concept}.
Low fuel flow scenarios may occur during rapid deceleration when sudden changes in the throttle rapidly reduce the fuel flow rate while the rate of change of air flow rate is slower~\cite{rosfjord1995evaluation}.
Extreme low fuel conditions can also arise during the adoption of an ultra-lean premixed mode of combustion to avoid $NO_{X}$ formation.
On the other hand, to keep the power input constant, there can be a continuous increment of air flow rate leading to highly turbulent combustion and, thereafter, LBO~\cite{mondal2022early,thampi2015intermittent}.
LBO can also be caused by large variations in the fuel flow, which can occur when there are sudden changes in power requirement~\cite{meegaha}.

As the lean combustors are used in a broad spectrum of industrial applications such as ground-based engines~\cite{mcdonell2013ground}, aero-engines~\cite{mularz1979lean,palies2019lean} and other gas turbine engines~\cite{bahlmann1994development}, the possibility of LBO can have a significant impact on many such systems.
For the ground-based engines in the power sector, LBO can cause unexpected stoppage of the engine, collapsing the entire power network~~\cite{meegaha}.
Such disruptions of engine activity affect productivity and can cause significantly higher maintenance costs.
On the other hand, for aviation, the risk of LBO is even higher when there is a low fuel flow scenario \cite{rosfjord1995evaluation}, and it can be challenging to recover from such a sudden flame loss during the running condition.
Thus, the study of LBO has become of great importance to the science community when engineers want to deal with "clean energy" technology (e.g., lean combustion). Nevertheless, it is difficult to determine the LBO regimes during the design phase, and therefore, developing an online detection tool for LBO can be an important contribution to different applications.

Previous works in the context of LBO detection are mostly concentrated on the premixed type of combustion (i.e., fixed degree of air-fuel mixing) \cite{muruganandam2005active, prakash2007lean, muruganandam2006sensing, nair2005acoustic, depremixed}.
In aero-engines, fuel is injected close to the reaction chamber, and therefore, LBO detection is also crucial for partially premixed combustion processes, where the degree of fuel-air mixing varies~\cite{de2019investigation}. 
There has been a recent focus on LBO prediction in partially premixed combustion \cite{de2019investigation, de2019use, de2020application} and researchers have proposed different statistical~\cite{chaudhari2011investigation,muruganandam2006sensing}, flame emissions based~\cite{chaudhari2013flame,de2019use}, nonlinear dynamics~\cite{unni2016precursors,de2020application} and data-driven techniques \cite{mukhopadhyay2013lean, sarkar2015dynamic, dey2015cross, mondal2022early,de2021early}.
However, the online LBO detection frameworks should be accurate and computationally fast simultaneously.
Some studies have focused on the computational time aspect of LBO \cite{de2020application,mondal2022early,de2021early}.
But, further exploration is essential to simultaneously consider the aspects of detection accuracy, computation time, and robustness.
While the application of deep learning models to detect combustion instability has started recently \cite{sarkar2015early1, akintayo2016prognostics, gangopadhyay2020deep, gangopadhyay2020interpretable, gangopadhyay20213d,gangopadhyay2021cross}, there has been no research work till now on LBO detection using deep learning.

In the present work, we propose a deep learning approach to detect LBO in a robust manner. 
We formulate the problem in such a way that detection performance can be tested for robustness.
We compare the performance of the proposed framework against three baseline methods in terms of detection accuracy and computation time. The results demonstrate that the proposed framework is the most accurate, computationally fastest and robust to generalize well for test datasets from different protocols.

% It shows the potential to be implemented as an online LBO detector in engines to detect lean blowout condition early.
% Quasi-static time-series datasets are captured at different conditions starting from far from blowout till we reach blowout.

We summarize the contributions of this work as follows:

\begin{itemize}

    \item To the best of our knowledge, this is the first work using deep learning to detect lean blowout in combustion systems.
    
    \item Our proposed LBO detection framework is dependent on pressure time-series data, which is a much convenient way of collecting data from real world combustors. Our training scheme is facilitated by the use of domain knowledge based labeling technique, which is based on flame images.
    
    \item To show the robustness of our model we test in different conditions near the lean blowout regime and we use three different baseline models for comparison in terms of accuracy and computation time.
    
    \item Our proposed deep learning approach is more accurate, computationally faster and captures the transition to lean blowout regime better than the baseline methods. It shows the potential to be implemented as an online detector to prevent the adverse effects of lean blowout in combustion systems.
    
\end{itemize}

\section{Methods}

In this portion, we describe the details of the experimental setup (Section~\ref{Details of the Model Combustor}), dataset (Section~\ref{Dataset Collection}), and the domain knowledge-based label annotation technique (Section~\ref{Labels Annotated Using Domain Knowledge}).
The notations and problem formulation are described.
Thereafter, we present the details of our proposed deep model architecture (Sections~\ref{train_test} $\&$ \ref{LSTM}) and discuss the baseline methods (Section~\ref{Baseline methods}) used for comparison.

\subsection{Details of the Model Combustor}\label{Details of the Model Combustor}

\begin{figure*}
\centering
\includegraphics[width=12cm, keepaspectratio]{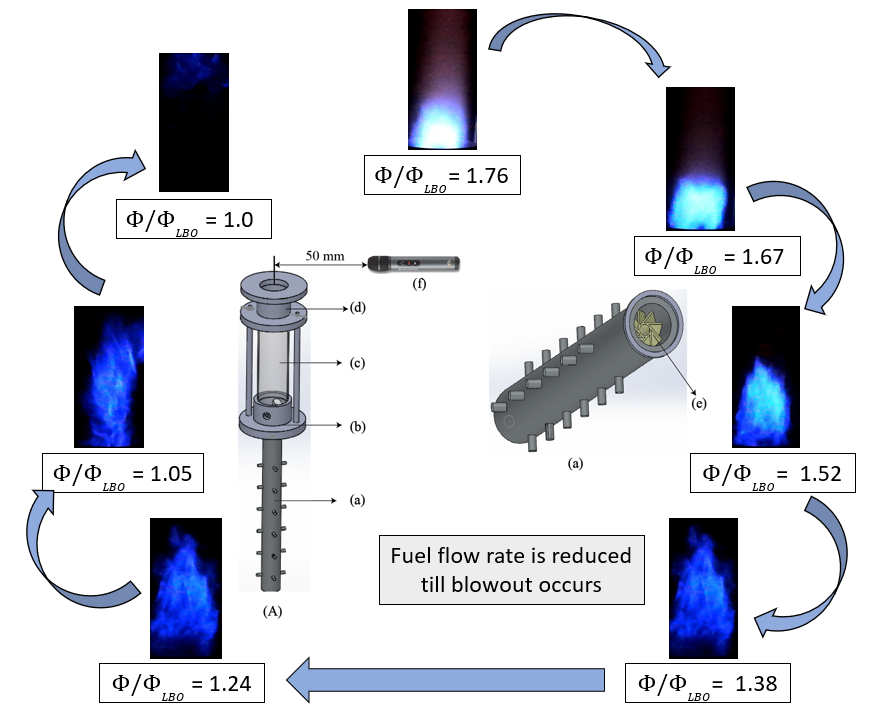}
\caption{The experimental setup is shown along with example images for different values of ${\Phi}/{\Phi_{LBO}}$. The fuel flow rate is reduced till blowout occurs at ${\Phi}/{\Phi_{LBO}} = 1$. The accessories of the experimental setup (A) are: (a) premixing chamber, (b) dump plane, (c) combustion chamber covered with quartz tube, (d) exhaust chamber and (e) swirler having 7 blades. Microphone (f), kept at a distance of 5 cm from the centre of the combustor rig, is used for capturing the timeseries data (combustion sound).}
\label{setup_lbo} 
\end{figure*}

The model combustor used in this study is the same as that has been used in \cite{de2019investigation, de2019use, de2020application}.
An illustration of the experimental setup is provided in Fig.~\ref{setup_lbo}. The experimental setup primarily consists of three major components: premixing chamber, combustion chamber, and exhaust chamber or outlet. The premixing chamber is constructed to control the quality of oxidizer-fuel mixing~\cite{de2019investigation}, and the chemical reaction takes place in the combustion chamber.

The premixing chamber, shown in Fig.~\ref{setup_lbo}(a), has the bottommost entry point for the oxidizer (air is used), and the other entry points from upstream to downstream of the premixer are used for the fuel. Therefore, the premixing chamber allows the mixing of fuel and oxidizer at different mixing lengths $L_F$ \cite{de2019investigation}.
At each entry point, there is a provision of four inlets (90$^{o}$ apart from each other) to reduce the asymmetry in the flow. 
After the topmost entry point, a swirler (swirl number = 1.26, the calculation has been shown in an earlier study \cite{de2019investigation}) is provided to improve the mixing, thereby enhancing the stability of combustion \cite{schefer2002combustion}.
The vane angle of the swirler is 60$^o$ to the axial direction. After the swirler, the sudden expansion in the flow area of the dump plane (Fig.\ref{setup_lbo}(b)) slows down the flow velocity and generates re-circulation regions, ensuring continuous feedback of heat radicals to the incoming mixture.
To provide optical accessibility, the combustion chamber (Fig.~\ref{setup_lbo}(c)) is surrounded by a quartz tube of length 200 mm and an outer diameter of 65 mm.
The burnt gases come out of the combustion chamber through the exhaust chamber.

\subsection{Dataset Collection}\label{Dataset Collection}

We have multiple protocols for dataset collection. The air flow rate is kept constant for each protocol, and the fuel flow rate is decreased to approach lean blowout. We vary the equivalence ratio ($\Phi$) from the stoichiometric condition (where $\Phi$ = 1) towards the blowout.
We measure the fuel and air flow rates using mass flow controllers (Alicat Scientific, MCR series, range of MFCs: 0-250 SLPM for fuel and 0-1500 SLPM for air). 
For each equivalence ratio, an audio reader (sampling frequency of 48 kHz) is used to record the acoustic time series, which captures the pressure fluctuations in the system. We also capture flame images utilized to understand the combustion behavior while annotating the acoustic time series for each experiment.

We collect data for five protocols with different air flow rates (70 SLPM, 75 SLPM, 80 SLPM, 85 SLPM, and 90 SLPM). We use the third fuel entry point (towards the downstream of the air entry point) as the fuel port, ensuring partial premixing of the fuel and oxidizer. For each of the five protocols, we keep decreasing the fuel flow rate and record acoustic pressure time series for different fuel flow rates. Therefore, there is a set of quasi-static datasets for each protocol. The equivalence ratio ($\Phi$) where lean blowout occurs is referred to as $\Phi_{LBO}$. We represent each quasi-static dataset by ${\Phi}/{\Phi_{LBO}}$ ratio where lean blowout implies ${\Phi}/{\Phi_{LBO}} = 1$ and higher values of ${\Phi}/{\Phi_{LBO}}$ indicate regimes away from lean blowout.

\subsection{Labels Annotated Using Domain Knowledge}\label{Labels Annotated Using Domain Knowledge}

The stability of a combustion process generally depends on factors like hydrodynamics of the flow~\cite{chaudhuri2008blowoff}, acoustic pressure fluctuations~\cite{domen2015detection} and heat release rate oscillations (influenced by chemical kinetics or combustion reaction rates)~\cite{de2019investigation}.
A discrepancy in these factors may lead to unusual behavior in the combustion zone. 
Domain experts can understand such transition of the combustion regime from the flame images, from which they can annotate each ${\Phi}/{\Phi_{LBO}}$ as healthy or unhealthy. 
In Fig.~\ref{setup_lbo}, sample flame images at different ${\Phi}/{\Phi_{LBO}}$ are shown to describe the annotation procedure.
Annotation `healthy' refers to conditions away from the LBO regime with ${\Phi}/{\Phi_{LBO}}$ values higher than the transition ${\Phi}/{\Phi_{LBO}}$. The conditions with ${\Phi}/{\Phi_{LBO}}$ lower than the transition ${\Phi}/{\Phi_{LBO}}$ are annotated `unhealthy' and are therefore near the LBO regime.

\begin{figure*}
\centering
\includegraphics[width=12cm, keepaspectratio]{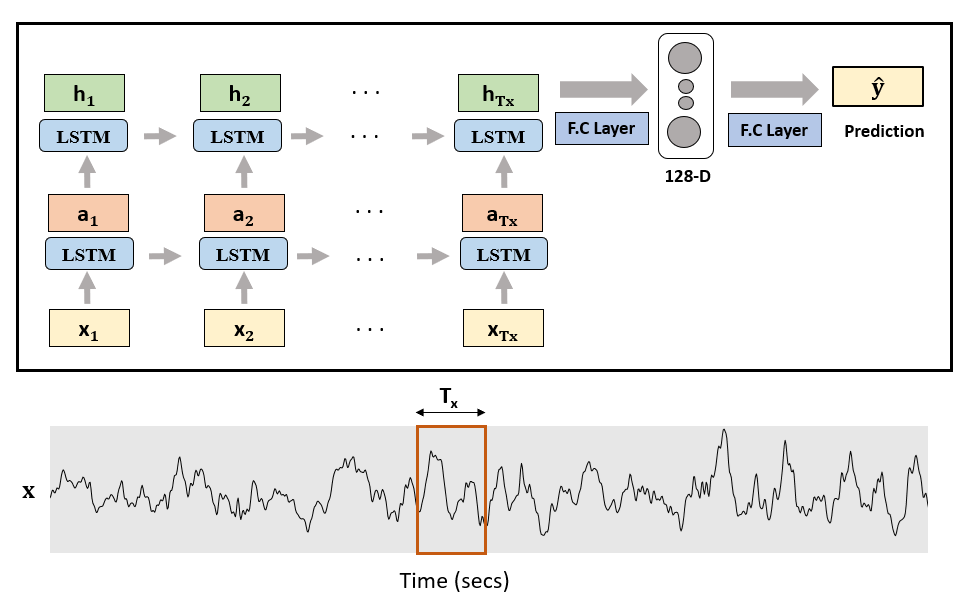}
\caption{Illustration of the proposed LSTM based deep learning model. Given the information for $T_x$ input time steps, the model learns to predict the immediate future time step $T_{x}+1$.}
\label{lstm_model_lbo} 
\end{figure*}

From the sample set of flame images shown in Fig.~\ref{setup_lbo}, near stoichiometric condition (at $\Phi/\Phi_{LBO}$ = 1.76), the combustion process looks stable as the flame adheres to the dump plane. This is observed when there is a strong equilibrium between the flame speed and the upcoming unburnt mixture \cite{turns1996introduction}. 
Therefore, flame exists at the inlet of the combustion chamber in the presence of high incoming flow ($Re \sim 3625$). 
The attachment exists till $\Phi/\Phi_{LBO} = 1.52$. A significant variation in combustion dynamics is observed at $\Phi/\Phi_{LBO} = 1.38$, where flame exists at a standoff distance from the chamber inlet. 
With the air flow rate fixed, as the fuel flow rate is decreased to lower values, the chemical reaction rate also decelerates which in turn reduces the heat release rate of combustion~\cite{turns1996introduction}.
As a result of this, the flame speed decreases impacting the attachment between the flame base and combustion inlet.
The bifurcation in the flame behavior at $\Phi/\Phi_{LBO} = 1.38$ can be attributed as transition~\cite{de2019investigation,de2020application}. 
Therefore, the flame behavior for $\Phi/\Phi_{LBO} > 1.38$ can be referred to as `healthy'.
With further reduction in ${\Phi}/{\Phi_{LBO}}$, the flame dynamics gets sufficiently weak and irregularities in behavior are noticed in $\Phi/\Phi_{LBO} = 1.24$ and $\Phi/\Phi_{LBO} = 1.05$ as demonstrated in Fig.~\ref{setup_lbo}.
As the condition approaches lean blowout, the rapid fluctuations start in the flame base with local extinction and re-ignition events \cite{de2019investigation, de2019use} ultimately leading to LBO, represented as $\Phi/\Phi_{LBO} = 1$.
The flame behaviors at $1 \le \Phi/\Phi_{LBO} < 1.38$ are referred to as `unhealthy' states.
The transition from healthy to unhealthy state can occur rapidly in real combustors and the probability of avoiding LBO becomes low once the flame enters into the critical regime (very close to LBO)~\cite{de2020application}. Therefore, it is preferable to predict the occurrence of LBO in early stage so that sufficient lead time is available to adopt precautionary measures.

\subsection{Training, Reference and Test Datasets}\label{train_test}

We keep one protocol (air flow rate = 90 SLPM) as the reference protocol and use the other four protocols as the test protocols. 

For the reference protocol, the details of the quasi-static datasets in terms of ${\Phi}/{\Phi_{LBO}}$ are provided below:

\begin{itemize}

    \item 90 SLPM: ${\Phi}/{\Phi_{LBO}}$ = [1, 1.142, 1.214, 1.285, 1.357, 1.428, 1.500, 1.571, 1.642, 1.714, 1.785]
    
\end{itemize}

The reason of choosing a protocol as `reference' is that we can find a transition state metric from the corresponding transition state which is already annotated. And, then, during testing, for any particular case, the computed metric can be compared against this transition state metric to directly predict that state as `healthy' or `unhealthy'.

The details of the quasi-static datasets in test protocols are:

\begin{itemize}

    \item 70 SLPM: ${\Phi}/{\Phi_{LBO}}$ = [1, 1.076, 1.153, 1.23, 1.307, 1.384, 1.461, 1.538, 1.615, 1.692, 1.769]
    
    \item 75 SLPM: ${\Phi}/{\Phi_{LBO}}$ = [1, 1.071, 1.143, 1.214, 1.285, 1.357, 1.428, 1.5, 1.571, 1.643, 1.714, 1.785]
    
    \item 80 SLPM: ${\Phi}/{\Phi_{LBO}}$ = [1, 1.066, 1.133, 1.2, 1.266, 1.33, 1.4, 1.466, 1.533, 1.6, 1.67]
    
    \item 85 SLPM: ${\Phi}/{\Phi_{LBO}}$ = [1, 1.071, 1.143, 1.214, 1.285, 1.357, 1.428, 1.5, 1.571, 1.643, 1.714, 1.785]
    
\end{itemize}

\subsection{LSTM Based Deep Learning Model}\label{LSTM}

We formulate the problem after introducing the notations that we use in the work. We denote the univariate time series input as $\mathbf{X}=[x_1,x_2,...,x_T]^{\top}\in\mathbb{R}^{T}$, where $T$ denotes the total length of the input time series.
We divide the entire dataset into consecutive time windows - each window of length $T_x$.
The consecutive windows can be defined as: $[x_1,...,x_{T_x}], [x_2,...,x_{({T_x}+1)}], ..., [x_{(T-{T_x}+1)},...,x_T]$.
We denote the $k$-th time window as
$\mathbf{x}^k=[x_k,...,x_{(k+{T_x}-1)}]^{\top}\in\mathbb{R}^{T_x}, k\in\{1,2,...,(T-{T_x}+1)\}$.
This pre-processing technique is followed for reference set and the test sets.
With a slight abuse of notation, the time series of the $k$-th time window can be denoted by $\mathbf{x}^k=[x^k_1,x^k_2,...,x^k_{T_x}]^{\top}\in\mathbb{R}^{T_x}$.

Recurrent Neural Network (RNN) based models have proved to be effective in different applications including machine translation \cite{cho2014learning}, speech recognition \cite{miao2015eesen}, performance monitoring of engineering systems \cite{gangopadhyay2020deep}, financial markets \cite{selvin2017stock} and healthcare \cite{jagannatha2016bidirectional}.
Recurrent Neural Networks (RNNs) are efficient in learning the temporal dependencies from time series data for accurate forecasting \cite{hewamalage2020recurrent}.
RNNs are trained using the error backpropagation algorithm in which the error gradients are propagated through the unrolled temporal layers.
In an RNN, at each time-step, the input is combined with the hidden state vector using a learned function to produce a new state vector.
The architecture of a RNN allows the information from the previous time-steps to persist by passing the hidden state from one step of the network to the other.
To overcome the problem of vanishing gradients of RNNs for long sequences \cite{bengio1994learning, gers1999learning}, an effective RNN architecture is Long Short Term Memory (LSTM) \cite{hochreiter1997long}. In an LSTM block, the input, output, and forget gates regulate the addition of any information to the cell state.
% Next, we provide the details of an LSTM block.

In this work, we develop an LSTM based deep learning model illustrated in Fig.~\ref{lstm_model_lbo}.
The first LSTM layer of the model reads the input sequence in order from $x_1$ to $x_{T_x}$ to compute the hidden states $[\mathbf{a}_{1},\mathbf{a}_{2},...,\mathbf{a}_{T_x}]$. This sequence of hidden states acts as the input to the second LSTM layer which generates the hidden states $[\mathbf{h_{1}}, \mathbf{h_{2}}, \cdots, \mathbf{h_{T_{x}}}]$. The state $\mathbf{h_{T_{x}}}$, capturing the summarized temporal information of the input time series, is passed through two fully connected layers to predict $\hat{\mathbf{y}}$.

We train the LSTM model on the ${\Phi}/{\Phi_{LBO}}=1$ dataset from the reference protocol using a training size of 90\%. Next, we randomly split the training set into training and validation sets with a validation size of 20\%. For each time series, the inputs are scaled to (0, 1). We perform experiments with different values of hyper-parameters to finalize the optimal set of hyper-parameters based on the validation set performance.
We use Adam optimizer (\cite{kingma2014adam}) with a learning rate of 0.001.
We train the model on an NVIDIA Titan RTX GPU for 50 epochs with a batch size of 512.

\subsection{Baseline Methods}\label{Baseline methods}

\textbf{Recurrent Neural Network (RNN):}
We use a deep learning baseline model based on RNN. To develop this model, we replace the LSTM units in Fig.~\ref{lstm_model_lbo} with RNN units keeping the architecture same. The RNN model is shown in supplementary materials.
The details of the RNN baseline model are provided in supplementary materials. 
We train the RNN model in a similar way as the LSTM model.

\begin{figure}
\centering
\includegraphics[width=9cm, keepaspectratio]{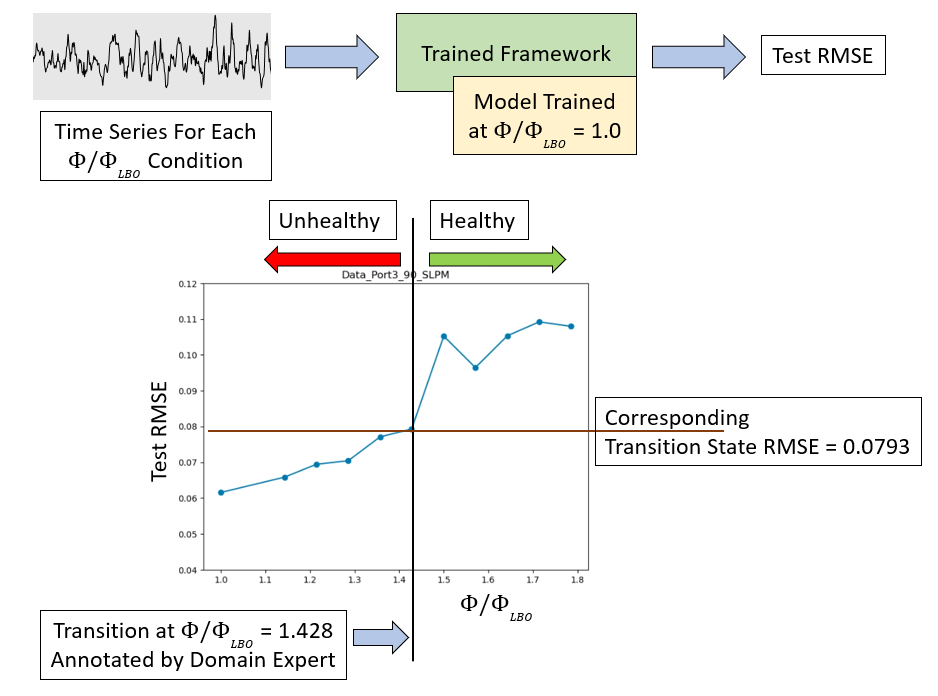}
\caption{Reference protocol results from the model LSTM. The test RMSE is 0.0793 corresponding to the transition ${\Phi}/{\Phi_{LBO}}=1.428$.}
\label{reference_LSTM} 
\end{figure}

\textbf{Hidden Markov Model (HMM):}
We use a machine learning modeling approach involving the Hidden Markov Model as one of the baseline methods.
Recently, Hidden Markov Models (HMMs) have been utilized for early prediction of a lean blowout from chemiluminescence time series data \cite{mondal2022early}.
HMMs, probabilistic frameworks for recognizing patterns in stochastic processes, have proved to be successful in speech recognition problems \cite{rabiner1989tutorial}. 
In an HMM, a sequence of observations is generated from a sequence of internal states. The likelihood of the observations depends on the states which are hidden. It is assumed that the transitions between the hidden states follow a first-order Markov chain.
An HMM is defined by the start probability vector $\mathbf{\Pi}$, transition probability matrix $\mathbf{A}$, and emission probability of an observation $\mathbf{\Theta}_i$.

In the training process, we utilize the ${\Phi}/{\Phi_{LBO}}=1$ dataset from the reference protocol using the same training size (90\%) as used in the case of the LSTM model.
The number of states which give the least Bayesian Information Criterion is chosen as the optimal number of states.
After getting the trained optimal HMM, we divide the training dataset into consecutive time windows of length $T_x$ to compute the log-likelihood for each window. 

To perform prediction for a test time window $k$, we first predict the log-likelihood using the trained HMM and then compare that with the log-likelihoods of the training set windows to find the closest window $j$. That window $j$ can be considered most similar to the test time window $k$.
The predicted change is computed as $\hat{\mathbf{y}}_{j} - \mathbf{x}^j_{T_x}$ and then it is added to $\mathbf{x}^k_{T_x}$ to get $\hat{\mathbf{y}}_{k}$.

\begin{figure*}
\centering
\includegraphics[width=18cm, keepaspectratio]{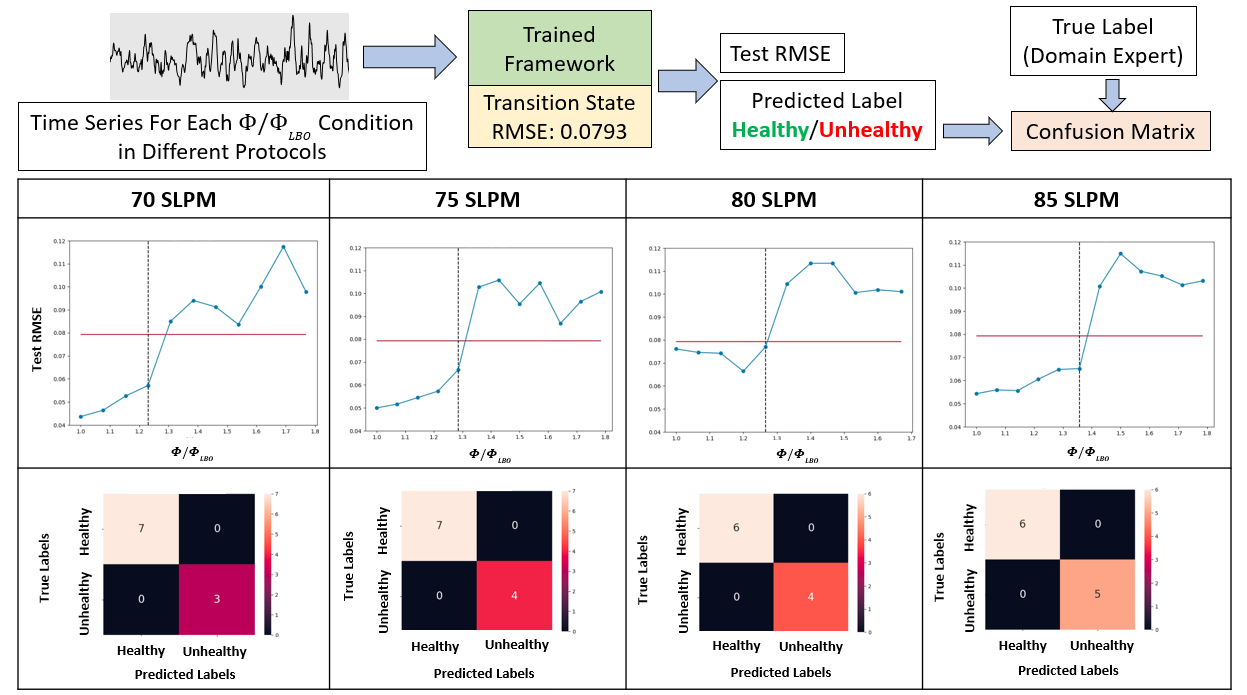}
\caption{Results for the quasi-static datasets from the four test protocols using the LSTM model. Test RMSE plot and confusion matrix are shown for each test protocol.}
\label{test_LSTM} 
\end{figure*}

\textbf{Translational Error:}
As another baseline method, we use an existing and well-known method -  translational error \cite{gotoda2014detection}. Translational error has been proposed as a suitable candidate for prediction and control of lean blowout \cite{gotoda2014detection}.
It is a nonlinear dynamic tool capable of identifying the dissimilarity in the directions of neighboring trajectories. Mathematically, it is expressed as:     

\begin{equation}
    E_{trans}=  \frac{1}{K+1}\sum_{k=0}^K \frac{||v(t_k)-\bar{v}||^2}{||\bar{v}||^2},
\label{Etrans}
\end{equation}

where, $\bar{v} =  \frac{1}{K+1} \sum_{k=0}^K v(t_k )$. Here, $v(t_k) = {P}^\prime(t_k+\tau_d)-{P}^\prime(t_k$). In Eq.\ref{Etrans}, $K$ indicates the number of neighbouring points of ${P}^\prime(t_i)$. The values of $K$ is considered here as 5 as suggested in the previous study \cite{gotoda2014detection}. 

The median of $E_{trans}$ values is estimated for 100 randomly chosen phase space vectors, ${P}^\prime(t_i)$. 
Therefore, the computation of translational error has a randomness factor in it, for which we use three independent computations in the case of the test dataset to report the mean and standard deviation. 
And to compare the computation time with other methods, we use the mean computation time from the three computations.
There are two other important parameters used in constructing the phase space vectors: time delay and optimum embedding dimension. These are derived after experiments with the training dataset ${\Phi}/{\Phi_{LBO}}=1$ from the reference protocol.

\begin{figure}
\centering
\includegraphics[width=8cm,keepaspectratio]{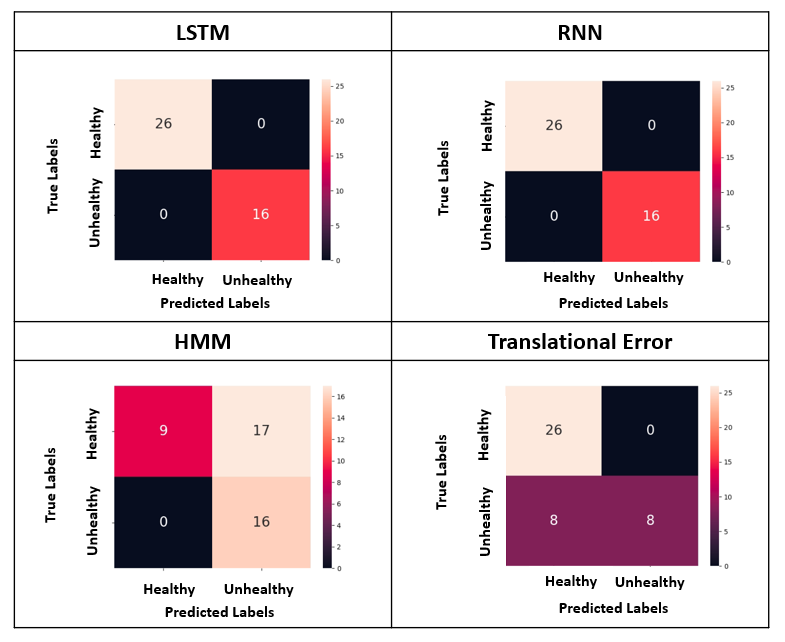}
\caption{Model-wise confusion matrices considering all the test protocols.}
\label{conf_matrix_lbo} 
\end{figure}

\section{Results}

\subsection{Transition State Error From Reference Set}

We utilize the reference protocol (90 SLPM) to get the transition state metric. As a metric, we use root mean square error (RMSE), which is computed between the predicted output and target output of the test set samples.
The reference protocol results for the LSTM model are shown in Fig.~\ref{reference_LSTM}.

We have the model trained at ${\Phi}/{\Phi_{LBO}} = 1$ and test the model on the unseen 10\% of the dataset to get the test RMSE at ${\Phi}/{\Phi_{LBO}}=1$. Next, we test the model on the quasi-static datasets at  ${\Phi}/{\Phi_{LBO}}=[1.142, 1.214, 1.285, 1.357, 1.428, 1.500, 1.571, 1.642, 1.714, \\ 1.785]$. For time series at each ${\Phi}/{\Phi_{LBO}}$ ratio, we get a corresponding test RMSE. 

From Fig.~\ref{reference_LSTM}, as the model has been trained on ${\Phi}/{\Phi_{LBO}} = 1$, we get the lowest test RMSE at ${\Phi}/{\Phi_{LBO}} = 1$ and with increasing ${\Phi}/{\Phi_{LBO}}$, the test RMSE increases. The transition state has been annotated at ${\Phi}/{\Phi_{LBO}} = 1.428$ by the domain experts, and the corresponding transition state RMSE is 0.0793. This information will be used to predict `healthy'/`unhealthy' labels while testing the LSTM model. Test RMSE values higher than 0.0793 can be labeled as `healthy' or away from the LBO regime, and test RMSE values lower than 0.0793 can be predicted as `unhealthy' or near the LBO regime.

The reference protocol results for the RNN model, HMM model and translational error method are provided in supplementary materials. For the translational error method, we perform three independent computations as the method has a randomness factor and compute the mean and standard deviation.

\begin{figure*}
\centering
\includegraphics[width=17cm, keepaspectratio]{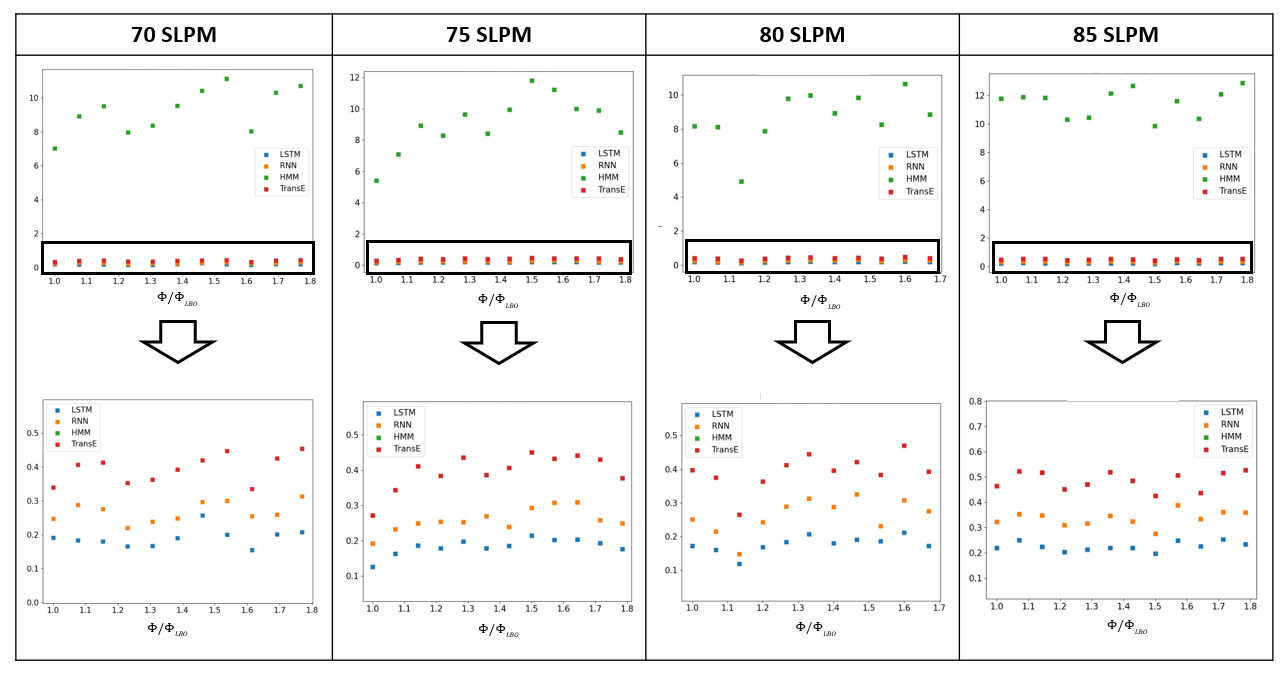}
\caption{Comparison of computation time for LSTM, RNN, HMM and Translational Error for different datasets in test protocols.}
\label{comp_time} 
\end{figure*}

\subsection{Performance on the Test Protocols}

We have four test protocols - 70 SLPM, 75 SLPM, 80 SLPM, 85 SLPM, and each test protocol consists of a set of quasi-static datasets.
For each quasi-static dataset in these protocols, the trained model is used to predict RMSE. The predicted label is `healthy' or `unhealthy' depending on whether the RMSE value is higher or lower than the transition state RMSE. 
Predicted labels are compared against actual labels for each test protocol to compute the confusion matrix.  

Results for the test protocols using the LSTM model are shown in Fig.~\ref{test_LSTM}. We have marked the transition state RMSE (0.0793) with a red line in each RMSE plot. The black dotted line in each plot denotes the actual transition state annotated by domain experts. The plots show a decreasing trend as we move closer to lean blowout, which occurs at ${\Phi}/{\Phi_{LBO}} = 1$. The LSTM model-based detection framework shows 100\% accuracy for all four protocols, as illustrated in the confusion matrices.

The test results for RNN model, HMM model and translational error method are provided in the supplementary materials. Like the LSTM model, the RNN model also demonstrates 100\% accuracy. Therefore, both the deep learning models can accurately detect the transition to lean blowout in combustion systems.
The HMM is not as effective as the deep learning models in capturing the transitions with changing conditions. 
The major disadvantage of the HMM model lies in false alarms, which can impact decision-making in real-time detection and control of LBO.
The randomness is one of the major disadvantages of using translational error as a detection tool.
Also, there are a lot of false negatives.
False-negative refers to falsely predicting negative or `healthy' while it is actually positive or `unhealthy.'
False negatives can be a serious issue in engine operation and lead to incorrect decision-making. 

% For all the test protocols, the confusion matrices show a lot of false positives, with the highest number of false positives for 75 SLPM, where it can only correctly identify one `healthy' condition.  Therefore, 

\subsection{Performance Comparison in Terms of Detection Accuracy and Computation Time}

To summarize the comparative performance considering all the test protocols, we demonstrate the overall confusion matrix for each method. The model-wise confusion matrices are shown in Fig.~\ref{conf_matrix_lbo}.

For both the LSTM and RNN models, the accuracy is 100\% for `healthy' and `unhealthy' datasets. From Fig.~\ref{conf_matrix_lbo}, HMM can detect the `unhealthy' datasets correctly, but it predicts most of the `healthy' datasets as `unhealthy,' leading to a lot of false positives. False positives refer to falsely predicting `healthy' conditions as `unhealthy' or positive. Out of 16 `healthy' conditions, it predicts 9 conditions as `healthy' and the remaining 17 as `unhealthy.'
Using the method of translational error, the detection accuracy for `healthy' is 100\%. But, it falsely predicts half of the `unhealthy' conditions as `healthy,' resulting in false negatives. It is highly important to have a detection framework with low false negatives to prevent the adverse effects of a lean blowout in combustion systems. Overall, the deep learning frameworks (LSTM, RNN) outperform Hidden Markov Model and Translational Error.

Next, we compare the computation time of the methods for each ${\Phi}/{\Phi_{LBO}}$ in the test protocols. We use the same computation (CPU) resources for an unbiased comparison, and the results are shown in Fig.~\ref{comp_time}. We observe that computation times of HMM are much higher than that of LSTM, RNN, and Translational Error. The computation time is the least for our proposed LSTM-based deep learning framework, followed by the RNN model and translational error.
Therefore, in terms of both detection accuracy and computation time, our proposed LSTM model outperforms the baseline methods and can therefore has the potential to be utilized for real-time detection and control of lean blowout. 

\section{Conclusion}

The adaption of lean combustion technology helps to reduce the $NO_{X}$ emissions, although it often leads to the occurrence of lean blowout due to the extreme leanness of the fuel-air mixture. Such an extinction of flame can lead to sudden loss of power in combustion systems causing unwanted adverse events. 
Therefore, It is essential to develop effective and feasible frameworks for real-time accurate detection and control of LBO.
With this motivation, in this paper, we develop a deep learning-based framework for detecting lean blowout in combustion systems.

In the present work, we take root mean square error (RMSE) as a metric to identify the combustion state (healthy or unhealthy). The proposed framework based on Long Short Term Memory is robust enough to detect the transition to the LBO regime in unseen test protocols. Further, we similarly compute metric using three baseline models: RNN, HMM, and Translational error. We find that our proposed framework is more accurate and computationally faster than these baseline methods across different test protocols. As the first deep learning work on LBO detection, it can be used for real-time performance monitoring in engines. A similar modeling approach can be adopted to develop control mechanisms of other cyber-physical systems using time series as a sensing modality.

\section*{Supplementary Material}
See the supplementary material for reference protocol results of the baseline methods RNN, HMM and Translational Error.

\section*{Authors' Contributions}
A.M., S.Sen. and S.D. conceived the combustion experiment(s). S.D. conducted the combustion experiment(s) and collected the data. T.G. and S.S. have conceived the machine learning models. T.G. built the machine learning models and generated the machine learning results. T.G. and S.D. performed the analysis. T.G., S.D. and Q.L. generated the translational error results. T.G., S.D., A.M., S.Sen. and S.S. interpreted the results and contributed in writing the manuscript. All authors have contributed in reviewing the manuscript.

\section*{Acknowledgements}
This work has been supported in part by NSF grants CNS1954556 and CNS 1932033.

\bibliographystyle{elsarticle-num-names}
\bibliography{references.bib}

\vfill
\newpage

\section*{Supplementary Material}

\section*{Recurrent Neural Network (RNN) Baseline Model}
\begin{figure}[h]
\centering
\includegraphics[width=8cm, keepaspectratio]{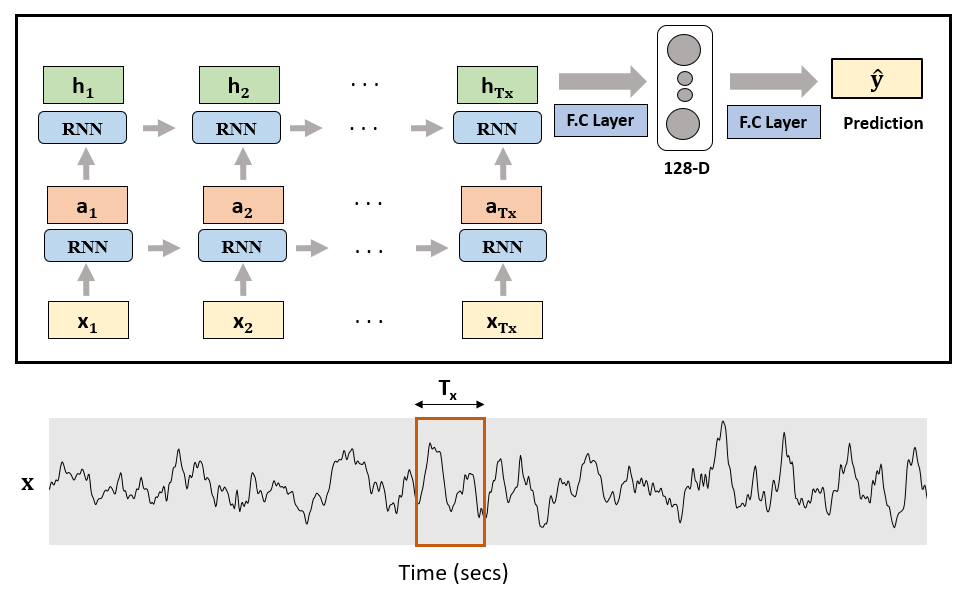}
\caption{Illustration of the baseline model based on RNN. Given the information for $T_x$ input time steps, the model learns to predict the immediate future time step $T_{x}+1$.}
\label{rnn_model_lbo} 
\end{figure}

We use a deep learning baseline model based on RNN, shown in Fig.~\ref{rnn_model_lbo}.
While describing the RNN based baseline model, to avoid clutter, we denote the input at time-step $t$ for a particular window $k$ as $x_t$ instead of $x^k_t$.
Therefore, for an univariate time-series, the input at $t$ is $x_{t}\in\mathbb{R}$.

The model comprises two stacked RNN layers and two fully connected layers.
We denote the hidden state corresponding to the first RNN layer with $\mathbf{a}_{*}$, where $*$ indicates the time-step. Similarly, the hidden state of the second layer of RNN is denoted as $\mathbf{h}_{*}$.
The hidden state of the first RNN layer at time-step $t-1$ is $\mathbf{a}_{t-1}\in\mathbb{R}^m$.

At time-step $t$, the input $x_t$ and previous hidden state $\mathbf{a}_{t-1}$ is passed to the RNN block, which computes $\mathbf{a}_{t}$ using a non-linear mapping as follows:
\begin{equation}\label{rnn_layer_1}
\mathbf{a}_{t} = tanh(W_{a}^{\top}[\mathbf{a}_{t-1};x_t] + b_{a})
\end{equation}
where $[\mathbf{a}_{t-1};x_t]\in\mathbb{R}^{m+1}$ with $\mathbf{a}_{t-1}\in\mathbb{R}^m$ the previous hidden state and $x_t\in\mathbb{R}$ the input for the $t$-th time step.
The parameters to learn are $W_{a}\in\mathbb{R}^{(m+1)\times m}$ and $b_{a}\in\mathbb{R}^{m}$.

The updated state $\mathbf{a}_{t}$ is passed onto the next time-step.
After reading the input sequence in order from $x_1$ to $x_{T_x}$, the first RNN layer returns a sequence of hidden states $\mathbf{A}=[\mathbf{a}_{1},\mathbf{a}_{2},...,\mathbf{a}_{T_x}]^{\top}$, where $\mathbf{a}_{t}\in\mathbb{R}^m$.
This sequence $\mathbf{A}$ acts as input to the second RNN layer. 

For the second RNN layer, the hidden state at time-step $t-1$ is denoted as $\mathbf{h}_{t-1}\in\mathbb{R}^n$.
At time-step $t$, the inputs to the RNN block are $\mathbf{h}_{t-1}\in\mathbb{R}^n$ and $\mathbf{a}_{t}\in\mathbb{R}^m$. The updated hidden state $\mathbf{h}_{t}\in\mathbb{R}^n$ is computed as:
\begin{equation}\label{rnn_layer_2}
\mathbf{h}_{t} = \textnormal{tanh}(W_{h}^{\top}[\mathbf{h}_{t-1};\mathbf{a}_{t}] + b_{h})
\end{equation}
where $[\mathbf{h}_{t-1};\mathbf{a}_{t}]\in\mathbb{R}^{n+m}$. The learnable parameters are $W_{h}\in\mathbb{R}^{(n+m)\times n}$ and $b_{h}\in\mathbb{R}^{n}$.

The output from the last time-step of the second RNN layer $\mathbf{h}_{T_x}\in\mathbb{R}^n$ is considered as the compressed information of the input sequence. 
Thereafter, one fully connected layer is used:
\begin{equation}\label{fc_layer}
\mathbf{d} = ReLU(W_{d}^{\top}\mathbf{h}_{T_x} + b_{d})
\end{equation}
The learnable parameters are $W_{d}\in\mathbb{R}^{n\times p}$ and $b_{d}\in\mathbb{R}^{p}$.

The prediction $\hat{\mathbf{y}}$ is computed after passing $\mathbf{d}\in\mathbb{R}^{p}$ 
through a fully connected layer.
\begin{equation}\label{linear_layer}
\hat{\mathbf{y}} = W_{y}^{\top}\mathbf{d} + b_{y}
\end{equation}
where the parameters to be learned are $W_{y}\in\mathbb{R}^p$ and $b_{y}\in\mathbb{R}$.

\section*{Results}

\subsection*{Transition State Error From Reference Set}

\begin{figure*}
\centering
\includegraphics[width=12cm, keepaspectratio]{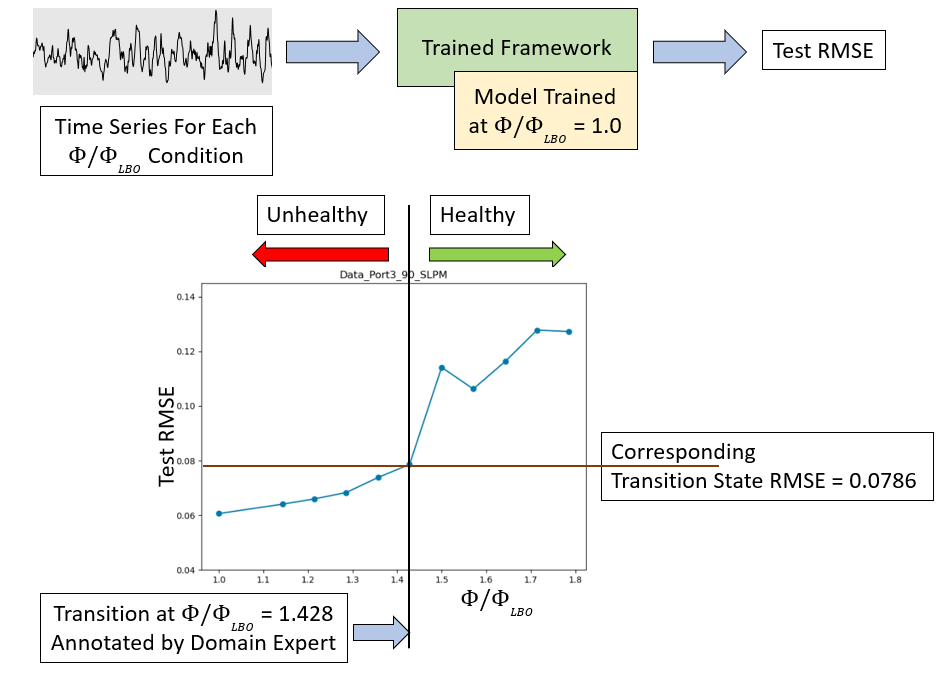}
\caption{Reference protocol results from the model RNN. The test RMSE is 0.0786 corresponding to the transition ${\Phi}/{\Phi_{LBO}}=1.428$.}
\label{reference_RNN} 
\end{figure*}

We demonstrate the reference protocol results for the RNN model in Fig.~\ref{reference_RNN}. 
We observe a similar trend as we observed for the LSTM model - with increasing ${\Phi}/{\Phi_{LBO}}$ the test RMSE increases, and the RMSE corresponding to the transition state is denoted as 0.0786.

\begin{figure*}
\centering
\includegraphics[width=12cm, keepaspectratio]{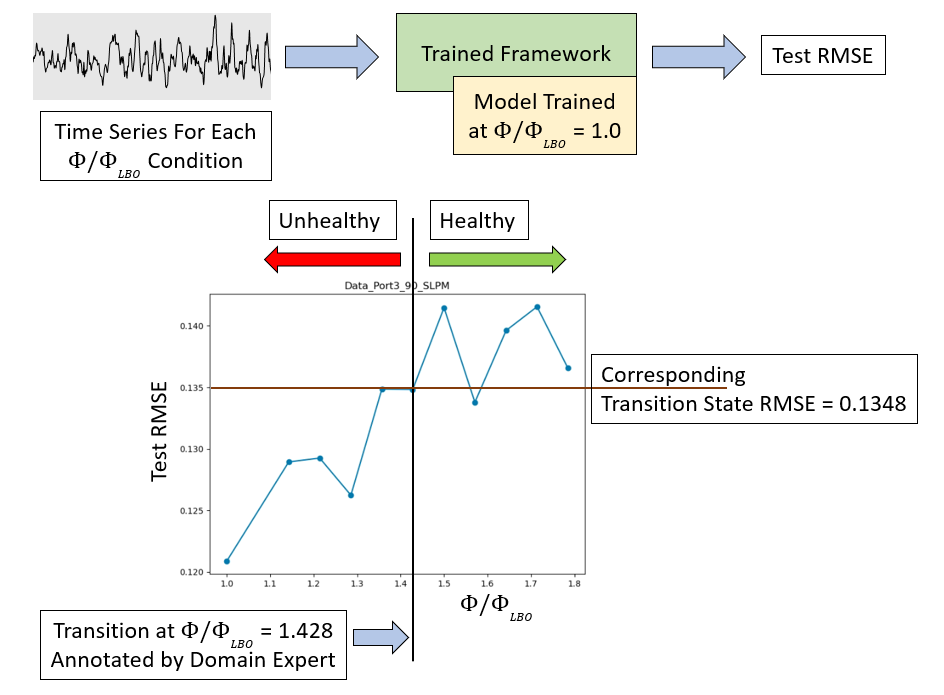}
\caption{Reference protocol results from the HMM. The test RMSE is 0.1348 corresponding to the transition ${\Phi}/{\Phi_{LBO}}=1.428$.}
\label{reference_HMM} 
\end{figure*}

The reference protocol results for the HMM are shown in Fig.~\ref{reference_HMM}.
Unlike the deep learning results of RNN and LSTM, we observe that the trend is not smooth in the case of the HMM. From Fig.~\ref{reference_HMM}, the transition state RMSE is 0.1348.

\begin{figure*}
\centering
\includegraphics[width=12cm, keepaspectratio]{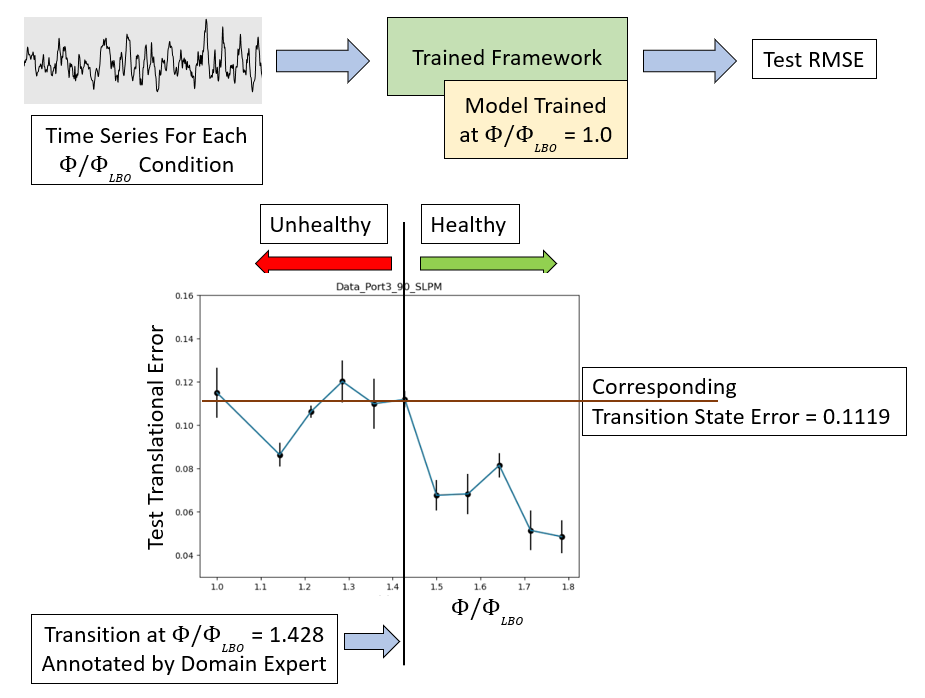}
\caption{Reference protocol results from the method translational error. The mean translational error is 0.1119 corresponding to the transition ${\Phi}/{\Phi_{LBO}}=1.428$.}
\label{reference_TransE} 
\end{figure*}

To get the reference protocol results for the method translational error, we perform three independent computations as the method has a randomness factor. We present the mean and standard deviation with error bars in Fig.~\ref{reference_TransE}.
We observe an increasing value of translational error as we go towards the lean blowout. The highest value of translational error is observed at ${\Phi}/{\Phi_{LBO}} = 1$. The mean translational error corresponding to the transition state is 0.1119. This error value will be used to predict the labels for the quasi-static datasets in the test protocols.

\subsection*{Performance on the Test Protocols}

\begin{figure*}
\centering
\includegraphics[width=16.5cm,keepaspectratio]{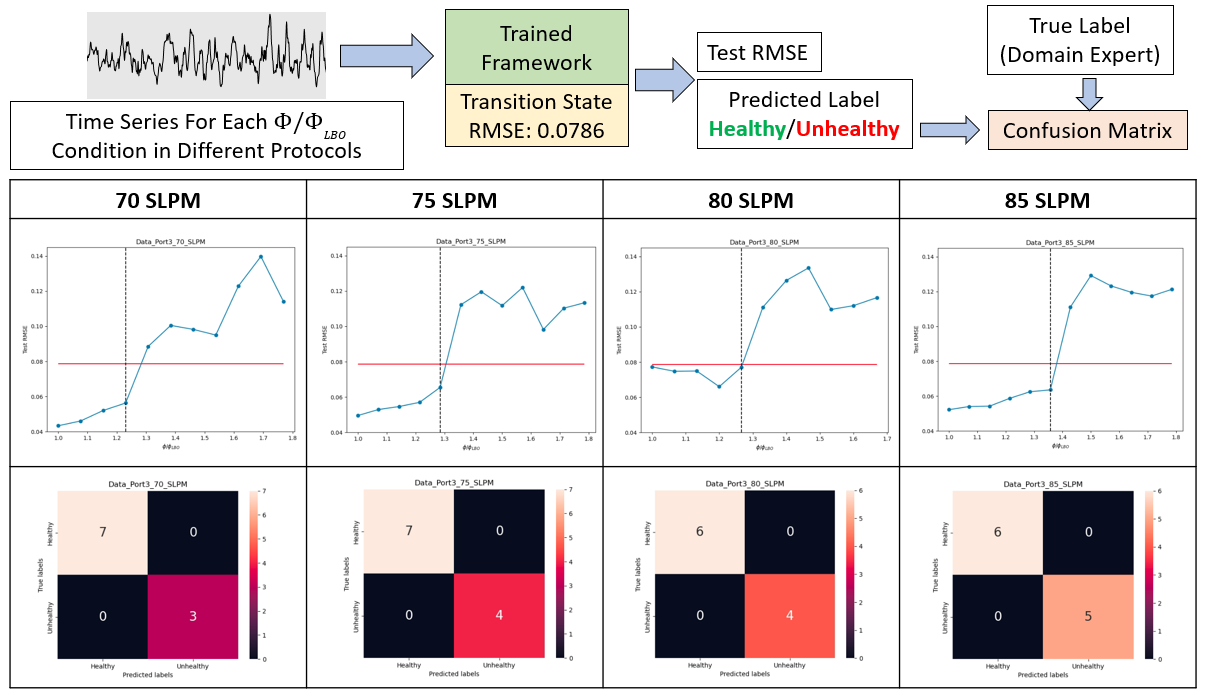}
\caption{Results for the quasi-static datasets from the four test protocols using the RNN model. Test RMSE plot and confusion matrix are shown for each test protocol.}
\label{test_RNN} 
\end{figure*}

We present the performance of the RNN model for the test protocols in Fig.~\ref{test_RNN}.
For each quasi-static dataset in the test protocols, the trained model predicts an RMSE, and the RMSE values are plotted with varying ${\Phi}/{\Phi_{LBO}}$. The confusion matrices are computed using the actual labels to evaluate the performance. Like the LSTM model, the RNN model also demonstrates 100\% accuracy. Therefore, both the deep learning models can accurately detect the transition to lean blowout in combustion systems.

\begin{figure*}
\centering
\includegraphics[width=16.5cm, keepaspectratio]{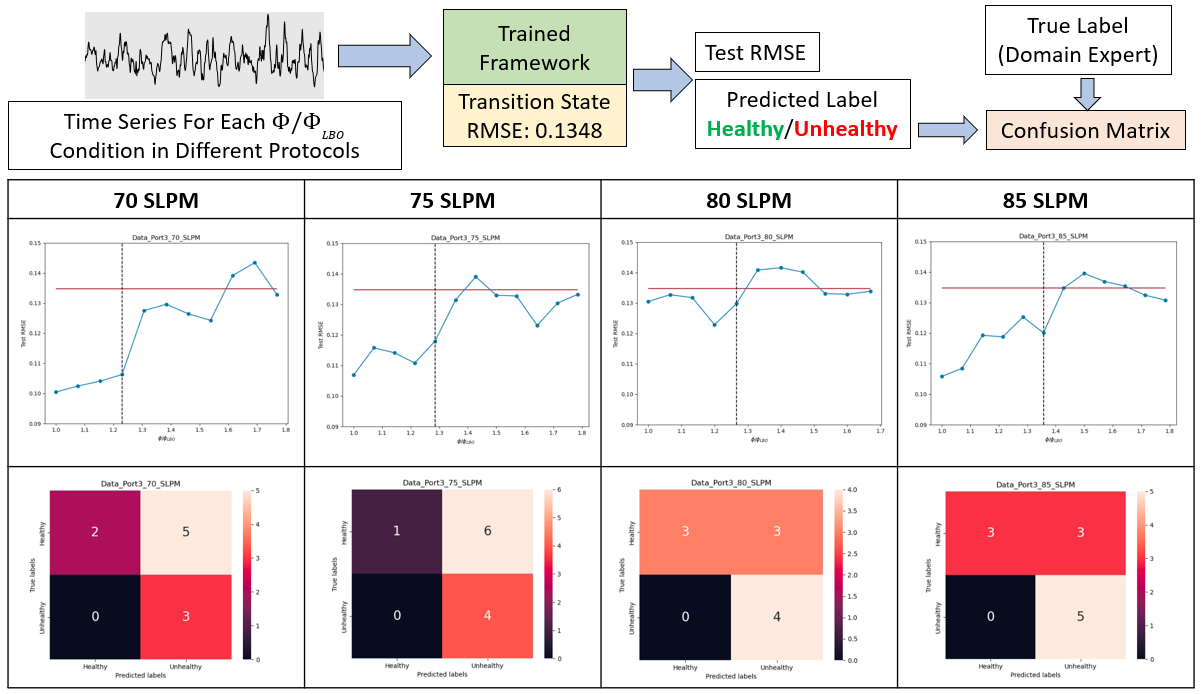}
\caption{Results for the quasi-static datasets from the four test protocols using the HMM. Test RMSE plot and confusion matrix are shown for each test protocol.}
\label{test_HMM}
\end{figure*}

Test results using the Hidden Markov Model are shown in Fig.~\ref{test_HMM}. The RMSE plots show that HMM is not as effective as the deep learning models in capturing the transitions with changing conditions. For all the test protocols, the confusion matrices show a lot of false positives, with the highest number of false positives for 75 SLPM, where it can only correctly identify one `healthy' condition.  Therefore, the major disadvantage of the HMM model lies in false alarms, which can impact decision-making in real-time detection and control of LBO.

\begin{figure*}
\centering
\includegraphics[width=16.5cm,keepaspectratio]{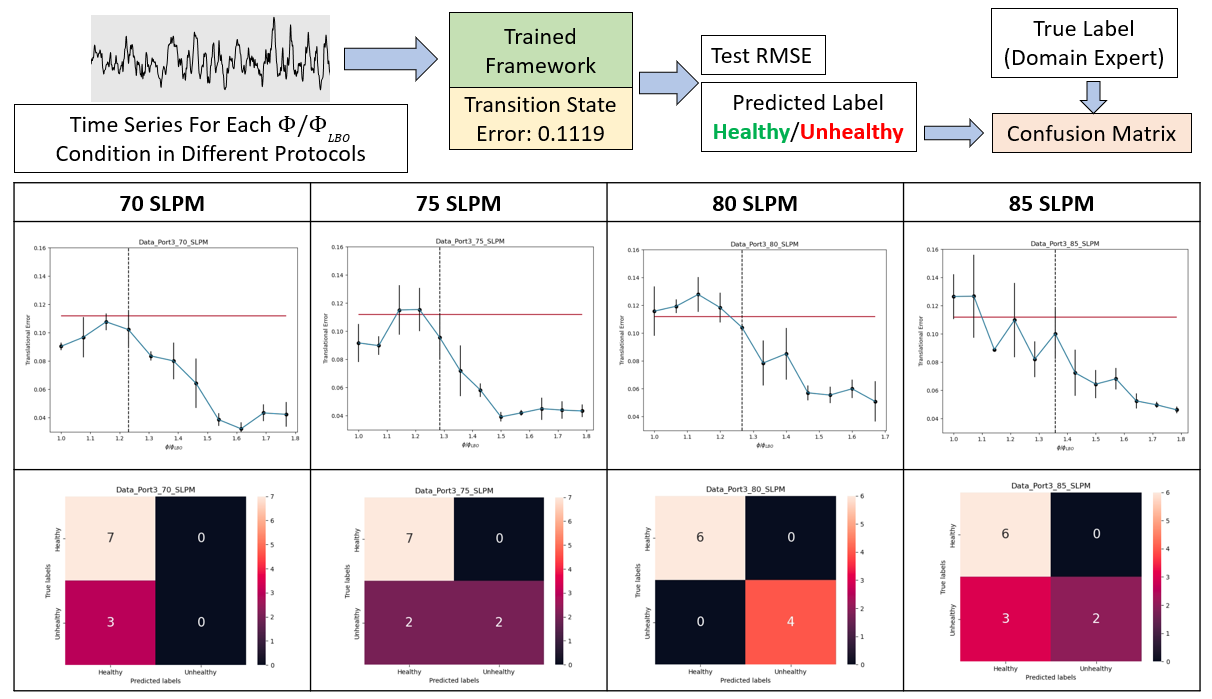}
\caption{Results for the quasi-static datasets from the four test protocols using the translational error method. Test RMSE plot and confusion matrix are shown for each test protocol.}
\label{test_TransE} 
\end{figure*}

We demonstrate the test results using the translational error method in Fig.~\ref{test_TransE}. 
While the mean and standard deviation of the error have been shown in the plot for each ${\Phi}/{\Phi_{LBO}}$, the confusion matrix computation is based on the mean translational error value. 
From the plots, we observe that the standard deviation values are pretty high for some datasets. This randomness is one of the major disadvantages of using translational error as a detection tool. While it can be a useful tool for offline analysis of lean blowout, the deep learning models outperform this method, as evident from the confusion matrices of Fig.~\ref{test_TransE}. Except for 80 SLPM, for all the other test protocols, there are a lot of false negatives.
False-negative refers to falsely predicting negative or `healthy' while it is actually positive or `unhealthy.'
False negatives can be a serious issue in engine operation and lead to incorrect decision-making. 


\end{document}